\PassOptionsToPackage{table}{xcolor}
\documentclass{ifm-tr}

\usepackage{amsmath,amsfonts,bm}

\def\eqref#1{equation~\ref{#1}}

\def\1{\bm{1}}

\DeclareMathAlphabet{\mathsfit}{\encodingdefault}{\sfdefault}{m}{sl}
\SetMathAlphabet{\mathsfit}{bold}{\encodingdefault}{\sfdefault}{bx}{n}

\usepackage{amsmath,amsfonts,amssymb,mathtools,bm}
\usepackage{amsthm}
\usepackage{thmtools}
\usepackage{algorithm}
\usepackage{algpseudocode}
\usepackage{enumitem}
\usepackage{siunitx}
\usepackage{tikz}
\usetikzlibrary{arrows.meta,fit,backgrounds,positioning}
\usepackage[utf8]{inputenc}
\usepackage{url}
\usepackage{xurl}
\usepackage{booktabs}
\usepackage{nicefrac}
\usepackage{graphicx}
\usepackage{tabularx}
\usepackage{xspace}
\usepackage{multirow}
\usepackage{makecell}
\usepackage{wrapfig}
\usepackage{lipsum}
\ifdefined\XeTeXrevision
  \providecommand{\faLightbulb}{}
\else
  \usepackage{fontawesome5}
\fi
\usepackage{xcolor}
\usepackage{listings}
\newcommand\rurl[1]{\href{https://#1}{\nolinkurl{#1}}}
\newcommand\grurl[1]{\href{https://github.com/#1}{\nolinkurl{#1}}}
\newcommand\hrurl[1]{\href{https://huggingface.co/#1}{\nolinkurl{#1}}}
\newcommand\wrurl[1]{\href{https://wandb.ai/#1}{\nolinkurl{#1}}}

\definecolor{colbg}{HTML}{F2F2F2}
\definecolor{dimrule}{HTML}{CCCCCC}
\definecolor{cellHighlight}{HTML}{dbefff}
\newcommand{\HL}{\cellcolor{cellHighlight}}

\definecolor{blue-primary}{HTML}{0081FB}
\definecolor{blue-light}{HTML}{D4ECFF}
\newtcolorbox{keytakeaways}[2][]{
    enhanced,
    colback=blue-light,
    colframe=blue-primary,
    colbacktitle=blue-primary,
    coltitle=white,
    fonttitle=\bfseries\large\color{white},
    title={\faLightbulb\, #2},
    after=\par\vspace{0.5em},
    arc=2.5pt,
    boxrule=0.6pt,
    left=9pt, right=9pt, top=8pt, bottom=8pt,
    drop fuzzy shadow,
    #1
}

\title{\bf GameScrub: Learning HUD Removal from Real Gameplay with Recovered Targets}

\author{\bf\color{black}Mohammad Areeb Qazi}\authorsep{,}
\author{\bf\color{black}Yiyan Hu}\authorsep{,}
\author{\bf\color{black}Yanan Wang}\authorsep{,}
\author{\bf\color{black}Xuezhi Liang}\authorsep{\\}
\author{\bf\color{black}Xin Li}\authorsep{,}
\author{\bf\color{black}Fan Yang}\authorsep{,}
\author{\bf\color{black}Xin Gao}\authorsep{,}
\author{\bf\color{black}Guangyi Liu}

\affiliation[]{Institute of Foundation Models}

\metadata[Project page]{\url{https://areeb2735.github.io/gamescrub/}}
\date{November 2026}

\abstract{
Gameplay videos are a rich source of experience for learning virtual-world dynamics, yet their frames contain heads-up displays (HUDs) that follow interface rules rather than scene dynamics. Removing HUDs would make this footage directly usable for learning world dynamics. However, training a removal model requires paired clips with and without HUDs, whereas real recordings contain only HUD-overlaid footage. Existing work synthesizes pairs by compositing HUD graphics onto HUD-free footage, requiring a separately curated clean corpus and producing HUDs that only approximate real ones. Models trained on such pairs leave 40 to 50\% of real HUD elements untouched. We instead recover pseudo-clean targets directly from real recordings by inpainting HUD regions with a pretrained video diffusion model. Preventing HUD traces from leaking into recovered targets is challenging because latent video autoencoders mix information across mask boundaries. Our training-free recovery pipeline, GameScrub-Data, pre-fills HUD regions with flow-guided background estimates before encoding, excludes HUDs from scene conditioning, and confines restoration to masked regions through pixel-space compositing. The resulting pairs train GameScrub, a video diffusion removal model that handles diverse HUDs without inference-time masks or clean reference images. On the in-the-wild Game2World benchmark, GameScrub achieves an artifact-adjusted removal score of 71.95, compared with 57.86 for the strongest reference-free GameCleaner configuration, and removes 90.06\% of annotated HUD elements versus 59.07\%. These results demonstrate that real recordings with recovered targets can support effective HUD removal, advancing HUD-free gameplay data for world models and controllable video generation.
}

\begin{document}
\maketitle





\section{Introduction}
\label{sec:intro}

\begin{figure}[htbp]
    \centering
    \includegraphics[width=\linewidth]{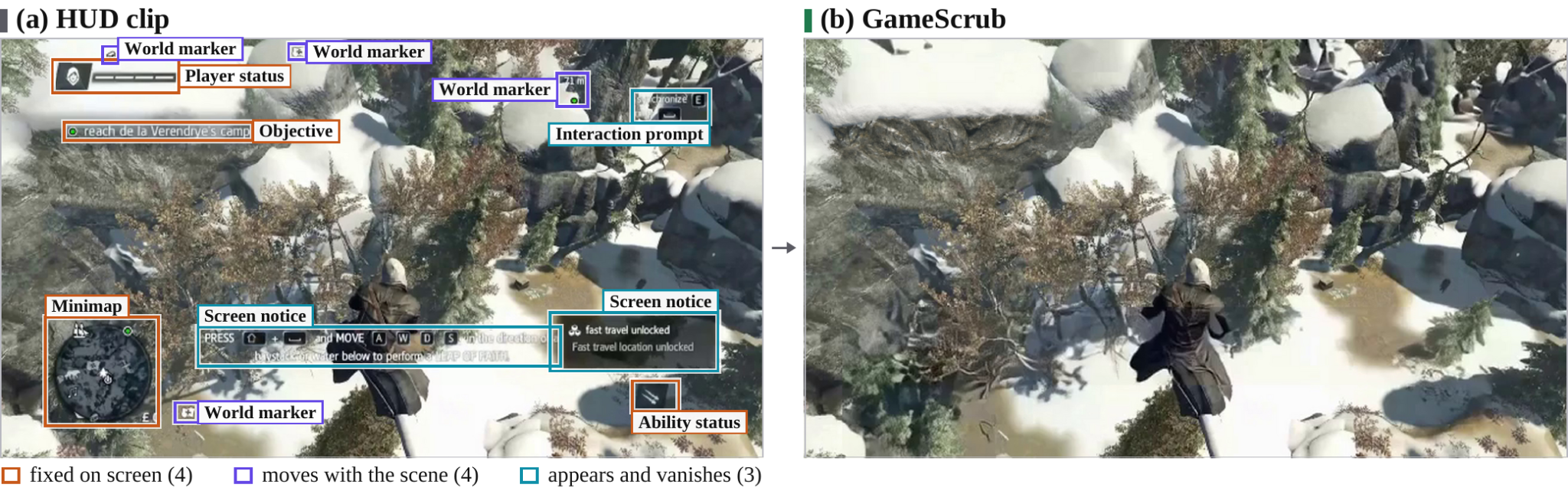}
    \caption{\textbf{GameScrub removes diverse HUDs from real gameplay.}
(a) A frame from the in-the-wild split of the Game2World benchmark~\cite{shen2026game2world} with its annotated HUD elements, colored by how they behave: fixed on screen, moving with the scene, or appearing and vanishing.
(b) The same frame of GameScrub's output: all the elements are removed.
Trained on real-input pairs, GameScrub removes HUDs in only 11 denoising steps without inference-time masks.}
    \label{fig:1}
\end{figure}

Gameplay videos capture two intertwined forms of visual dynamics: the evolution of virtual worlds and the behavior of the interfaces rendered over them. While the former provides valuable experience for learning interactive world models~\cite{valevski2025gamengen,alonso2024diamond,bruce2024genie,che2025gamegenx,zhang2025matrixgame,robbyant2026lingbot}, the latter introduces visual patterns that do not belong to the underlying scene. Heads-up displays (HUDs), such as minimaps, health bars, and crosshairs, are particularly pervasive in Internet gameplay recordings. They take many forms: some stay fixed on screen throughout a clip, such as minimaps and status bars; some move across it, such as markers that follow objects; and some appear and vanish with game events, such as notifications and subtitles. Their screen-space appearance and evolution follow interface conventions rather than the geometric and physical dynamics that world models aim to capture. As increasingly diverse Internet videos are used to train such models~\cite{he2026sekai2}, separating these two sources of visual information becomes important. Indeed, Game2World~\cite{shen2026game2world} demonstrates that removing HUDs from gameplay footage improves downstream video generation, by 6.83\% in overall VideoReward and 18.8\% in motion quality. A reliable HUD remover would therefore turn the large body of recorded gameplay into cleaner training data for world models and for controllable video generation, where an overlay that reacts to the player's actions is a pattern the model should not learn to reproduce.

Despite its importance, learning HUD removal from real gameplay is constrained by the absence of clean supervision. For videos recorded in the wild, the corresponding HUD-free frames are generally unavailable. Existing approaches address this problem by compositing synthetic HUDs onto clean gameplay, producing perfectly aligned training pairs. Game2World~\cite{shen2026game2world}, for example, constructs 96K such pairs to train GameCleaner. This route requires a separately curated corpus of HUD-free gameplay~\cite{shen2026game2world}, whereas ordinary recordings, which carry their HUDs, are far more plentiful. Also, synthetic composition can only approximate the diversity of authentic overlays, whose transparency, boundaries, animations, and event-dependent behaviors vary substantially across games. GameCleaner's reported artifact-adjusted removal drops from 95.36 on synthetic gameplay to 48.56 on real footage~\cite{shen2026game2world}, highlighting the difficulty of transferring removal models from synthesized to authentic observations. The failure has a specific form: on real clips GameCleaner leaves 40 to 50\% of the annotated HUD elements untouched and removes nearly all of the others cleanly (\cref{tab:main}), so it does not remove real HUDs badly, it fails to recognize them. This raises a question: must the clean target be exact if doing so requires approximating the very corruptions a model must learn to remove?

We explore the opposite trade-off. Rather than synthesizing corrupted inputs from clean videos, we recover pseudo-clean targets from authentic HUD-contaminated gameplay. This reverses the construction of training pairs: the input preserves the true appearance and dynamics of real HUDs, while the target is obtained through generative restoration. The resulting supervision is necessarily approximate, but it avoids introducing a synthetic approximation on the input side. Making this strategy effective, however, requires more than producing visually plausible clean videos. Any residual HUD or unintended scene alteration in the recovered targets becomes part of the supervision learned by the removal model. Reliable target recovery is therefore the main requirement for learning from real gameplay. No HUD-free footage is needed, any recorded gameplay can serve as input.

We address this challenge with GameScrub-Data, a training-free recovery pipeline, and GameScrub, a mask-free video editor trained on its output. A key obstacle in recovering reliable targets arises from the nonlocal representations of latent video diffusion models. Even when generation is restricted to a tight HUD mask, latent features outside that region may already encode HUD content through the autoencoder's spatial and temporal receptive fields. Preserving these features can retain unwanted artifacts, while expanding the mask unnecessarily exposes valid scene content to regeneration. GameScrub-Data addresses this trade-off before latent encoding: a flow-guided background estimate first replaces HUD-contaminated pixels, reducing their influence on the features used to condition restoration. A frozen video diffusion model then refines the missing content with masked latent denoising, while captions derived from HUD-masked observations prevent the text condition from reinforcing the overlays. Finally, pixel-space compositing retains the original video outside the HUD regions, confining the recovered supervision to the content that needs to change.

The resulting pairs allow GameScrub to learn HUD removal directly from authentic gameplay without requiring HUD masks as model inputs. The original video serves as an in-context reference for a pretrained video diffusion transformer adapted through low-rank updates, while the recovered clip provides the reconstruction target. Reference conditioning is maintained during both generation and subsequent refinement stages, allowing the model to recover occluded content while remaining grounded in the observed scene. This separates the requirements of supervision construction from those of deployment: HUD localization and iterative restoration are performed offline to recover training targets, whereas the learned editor operates without inference-time masks or supplied clean references. With a step-distilled backbone, GameScrub performs removal in 11 diffusion denoising steps.

\begin{figure*}[t]
  \centering
  \includegraphics[width=\linewidth]{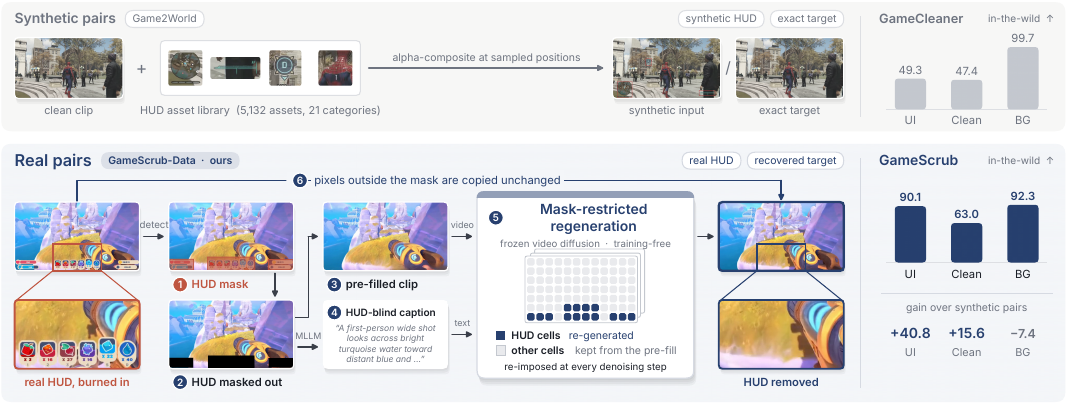}
  \caption{{\bf GameScrub-Data: Recovering paired supervision from real gameplay.} Top: Game2World~\cite{shen2026game2world} composites HUD assets onto clean footage, yielding synthetic inputs with exact targets. Bottom: GameScrub-Data retains real inputs and recovers pseudo-clean targets through HUD-aware conditioning, masked denoising, and pixel-space compositing. Right: system-level HUD removal (UI), artifact-free removal (Clean), and background preservation (BG) for standard GameCleaner and GameScrub on the in-the-wild benchmark (\cref{tab:main}).}
  \label{fig:teaser}
\end{figure*}

Experiments on the in-the-wild Game2World benchmark demonstrate the effectiveness of GameScrub on diverse real gameplay (see Figure~\ref{fig:1}). GameScrub achieves 71.95 artifact-adjusted removal, compared with 47.98 and 57.86 for the two GameCleaner configurations that take no reference image at inference, and a combined removal--preservation score of 82.13 against 78.88. The gain comes from recognizing HUDs: GameScrub removes 90.06\% of annotated elements where GameCleaner removes at most 59.07\%. Our contributions are summarized as follows:

\begin{itemize}
    \item \textbf{Contribution one.} We show that gameplay HUD removal can be learned from real HUD-contaminated recordings alone, with no HUD-free footage and no synthetic overlays, by recovering a pseudo-clean target for each recording.

    \item \textbf{Contribution two.} We develop GameScrub-Data, a training-free recovery pipeline that combines pre-encoding background filling, HUD-aware scene conditioning, and localized video diffusion to reduce contamination in recovered supervision.

    \item \textbf{Contribution three.} We present GameScrub, a reference-conditioned video diffusion editor learned from these pairs, enabling mask-free HUD removal on real gameplay with only 11 denoising steps.
\end{itemize}

\section{Related Work}
\label{sec:related}

\noindent{\bf Gameplay Videos and HUD Removal.}
Video world models learn environment dynamics from diverse visual experiences, ranging from Atari and DOOM to Minecraft and open-world games~\cite{alonso2024diamond,valevski2025gamengen,guo2025mineworld,che2025gamegenx,zhang2025matrixgame,robbyant2026lingbot}. Recent efforts further exploit large-scale Internet videos to improve the diversity of learned environments~\cite{bruce2024genie,he2026sekai2}. However, gameplay recordings contain heads-up displays (HUDs) that follow screen-space rules rather than the dynamics of the underlying world. Game2World~\cite{shen2026game2world} addresses this problem through a HUD taxonomy, a synthetic data engine, and a dedicated removal model. Its training pairs are constructed by overlaying synthetic HUDs onto clean gameplay, providing exact targets but introducing a discrepancy between synthesized and real HUD appearances. In contrast, GameScrub learns from authentic gameplay by recovering approximate clean targets while preserving the original HUD-contaminated inputs.

\noindent{\bf Video Inpainting and Latent Restoration.}
Video inpainting reconstructs missing content through motion-guided propagation~\cite{zeng2020sttn,li2022e2fgvi,zhou2023propainter} or generative priors~\cite{jiang2025vace,zi2025minimaxremover,fu2026effecterase}. Recent diffusion-based methods further improve object removal through background preservation or selective token processing~\cite{kushwaha2026objectwiper,wu2026yose}. However, HUD removal presents distinct challenges: overlays are screen-aligned, often semi-transparent, and may persist throughout a video, preventing direct observation of the underlying content at their screen locations. Preserving unmasked content is another challenge in latent diffusion. RePaint~\cite{lugmayr2022repaint} and Blended Latent Diffusion~\cite{avrahami2023blended} preserve source information through masked denoising, while recent studies reveal discrepancies between latent-space and pixel-space compositing caused by nonlocal autoencoder representations~\cite{bradbury2026latentmask,yao2026pixperfect,liu2026blend,wang2025enhanced}. Rather than correcting latent composition itself, GameScrub-Data addresses contamination before encoding: flow-guided pre-filling replaces HUD content with a background estimate, reducing its influence on the latent features intended to remain unchanged.

\noindent{\bf Mask-Free Video Editing and Generated Supervision.}
Instruction-guided video editors support mask-free content manipulation~\cite{lin2026kiwiedit,yu2026aurora,wu2026loomvideo}, while specialized methods address subtitle removal~\cite{clear2026,sedit2026}. However, game HUDs exhibit greater diversity in appearance, layout, transparency, and temporal behavior, making removal across unseen titles particularly challenging. Another line of work constructs paired supervision using pretrained generative models and automated annotation or filtering~\cite{bai2026ditto,huang2026ffp300k,li2026egoedit}. GameScrub follows this broader direction but adopts a different construction for HUD removal: instead of synthesizing overlays on clean footage, it recovers pseudo-clean targets from real HUD-contaminated videos. These pairs supervise a reference-conditioned video diffusion model that requires neither HUD masks nor externally supplied clean references at inference. Efficient adaptation is achieved using a step-distilled backbone~\cite{hacohen2026ltx2,ltx25} and low-rank tuning~\cite{hu2021lora,huang2024iclora}, enabling removal in 11 denoising steps.

\section{GameScrub-Data}
\label{sec:data}

\begin{figure}[t]
  \centering
  \includegraphics[width=\linewidth]{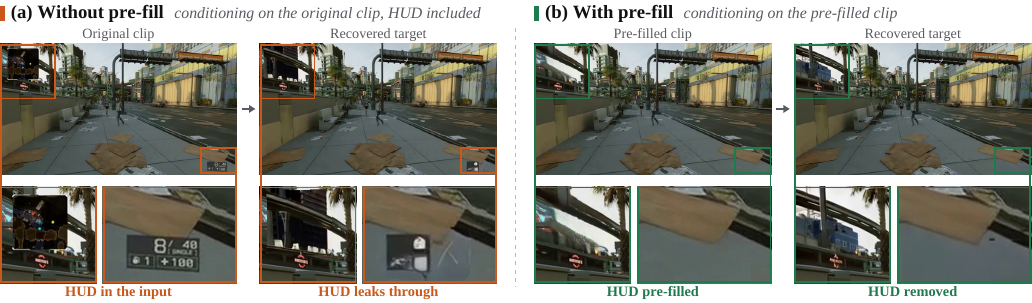}
  \caption{{\bf Pre-filling under a fixed tight mask.}
  On this clip, conditioning on the raw input leaves visible HUD remnants (top, orange), while replacing HUD pixels before encoding reduces these remnants (bottom, green).}
  \label{fig:leak}
\end{figure}

Real gameplay typically lacks aligned HUD-free targets.
GameScrub-Data constructs supervision by retaining each recorded clip as input and recovering a pseudo-clean target, rather than synthesizing the HUD corruption~\cite{shen2026game2world}. Recovery must limit both residual overlays and unintended scene changes, since either can become part of the training signal. We therefore suppress HUD cues in the restoration conditions, regenerate the masked content with a frozen video diffusion model, and preserve unmasked pixels through source compositing (see \cref{fig:teaser}). All components remain fixed during pair construction\footnote{Due to redistribution restrictions on the source gameplay footage, we will release the data construction pipeline rather than the dataset, enabling others to generate training pairs from their own videos.}.

\subsection{HUD Localization}
\label{sec:data-overview}
\label{sec:data-mask}

Let $x$ denote the input clip and $M_{\mathrm{px}}$ its per-frame binary HUD mask, obtained using a YOLO11n detector. The recovery model uses a spatially and temporally compressing autoencoder with encoder $\mathcal{E}$ and decoder $\mathcal{D}$. We construct the latent mask $M$ by marking each cell whose spatiotemporal pixel block overlaps the mask (see supplementary). Mask coverage affects supervision directly: missed HUD pixels remain in the target, whereas excess coverage exposes valid scene content to reconstruction.

\subsection{HUD-aware Conditioning}
\label{sec:data-conditioning}

A mask specifies where to edit, but HUD information can still enter through the conditions used for restoration.
We address the text and visual conditioning paths separately.

\noindent{\bf Scene caption.}
Captions of raw gameplay may describe HUD elements and encourage their regeneration. We instead provide an off-the-shelf multimodal language model with the HUD-masked clip, $x \odot (1-M_{\mathrm{px}})$, and use its scene description as the restoration prompt.
This hides detected overlays from the captioner rather than relying on it to distinguish them from scene content.

\noindent{\bf Background pre-fill.}
The encoder mixes information across spatial and temporal neighborhoods, so latent cells outside a tight mask may still encode HUD content~\cite{bradbury2026latentmask,yao2026pixperfect}.
Preserving these cells during denoising can retain overlay traces.
Mask dilation regenerates the affected neighborhood but also exposes valid background to unnecessary changes.

We instead replace the HUD before encoding. A flow-guided video inpainting model fills the masked region using surrounding background and neighboring frames, leaving unmasked pixels unchanged. The resulting clip $\tilde{x}$ changes what the encoder propagates across the mask boundary, rather than its receptive field.
This coarse fill supplies a visual condition, not the final target.
In \cref{fig:leak}, pre-filling reduces visible HUD residue under the same tight mask; removing it lowers the pipeline's AAR on the benchmark from 85.7 to 59.2, while the pre-fill alone, without the diffusion step, reaches 61.0 (see supplementary).

\subsection{Masked Target Recovery}
\label{sec:data-diffusion}

We encode $\tilde{x}$ and run the frozen video diffusion model conditioned on the scene caption.
Following~\cite{lugmayr2022repaint,avrahami2023blended}, after each denoising step, generated values are retained inside $M$, while the remaining cells are replaced with the pre-filled latents at the current noise level:
\begin{equation}
  z_t \leftarrow M \odot z_t
  + (1-M) \odot
  \big[(1-t)\,\mathcal{E}(\tilde{x}) + t\,\epsilon\big],
  \label{eq:snap}
\end{equation}
where $\epsilon$ is Gaussian noise, with $t=1$ denoting the noise endpoint and $t=0$ the clean endpoint.
This lets the masked latents evolve while anchoring the surrounding context to the pre-filled clip.
The final latents are decoded as $\hat{x}=\mathcal{D}(z_0)$.

\noindent{\bf Pixel-space compositing.}
Latent constraints alone do not ensure pixel-exact preservation after decoding~\cite{bradbury2026latentmask}.
We therefore construct the target as
\[
  y = M_{\mathrm{px}} \odot \hat{x}
    + (1-M_{\mathrm{px}}) \odot x.
\]
Each pair $(x,y)$ retains the authentic input, uses recovered content only inside the detected HUD mask, and copies the original pixels elsewhere.
The resulting supervision asks GameScrub to replace overlays while reproducing the observed scene outside them.
We evaluate the recovery stage with annotated HUD masks in \cref{sec:results-teacher}.

\section{GameScrub}
\label{sec:model}

\begin{figure*}[t]
  \centering
  \includegraphics[width=\linewidth]{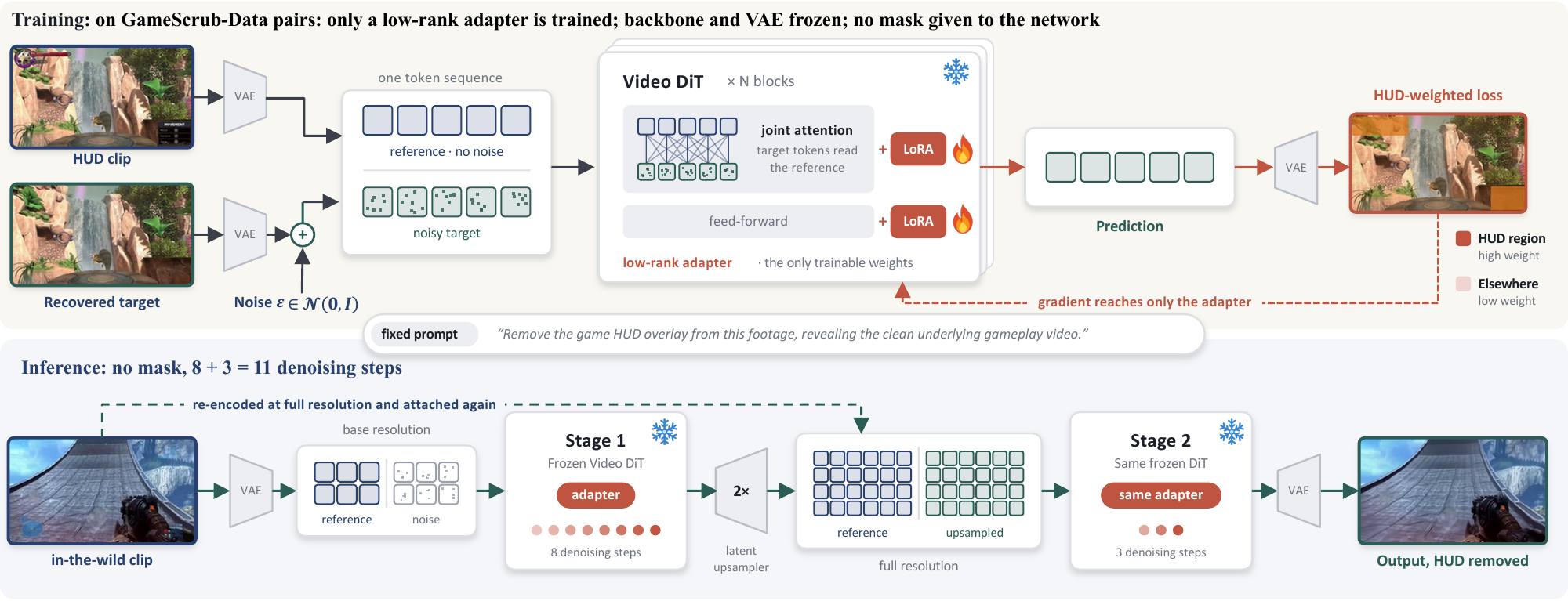}
  \caption{{\bf Overview of GameScrub.}
  Top: unnoised tokens from the HUD clip attend jointly with noisy target tokens. A target-only, HUD-weighted flow-matching loss trains the low-rank adapter while the backbone and autoencoder remain frozen. Bottom: the backbone's distilled sampler performs 8 base-resolution steps, frozen $2\times$ latent upsampling, and 3 full-resolution refinement steps. The HUD clip conditions both stages and is re-encoded for refinement, with the adapter active throughout.
  Training and inference use the same fixed instruction, without HUD masks as model inputs.}
  \label{fig:model}
\end{figure*}

GameScrub learns HUD removal from the recovered pairs, separating offline target construction from mask-free inference.
Unlike GameScrub-Data, it uses the original gameplay clip as its only visual input, without a HUD detector, clip-specific captioner, or background pre-fill at deployment.
The input clip provides scene context, while the recovered target specifies the desired edit (\cref{fig:model}).

\subsection{Reference-conditioned Adaptation}
\label{sec:model-arch}

We adapt a pretrained flow-matching video diffusion transformer using in-context LoRA~\cite{huang2024iclora,hu2021lora}.
The HUD clip and target are encoded by the same autoencoder.
Unnoised reference tokens $r$ are concatenated with noisy target tokens, allowing generation to attend directly to the observed scene through the backbone's self-attention (\cref{fig:model}, top).
Reference tokens remain at $t=0$ throughout.

LoRA modules on the attention and feed-forward projections are the only trainable parameters; the backbone and autoencoder remain frozen.
A single removal instruction $c$ is shared across all clips during training and inference, replacing the clip-specific scene captions used for target recovery.

\subsection{HUD-weighted Training}
\label{sec:model-prelim}
\label{sec:model-training}

For a training pair $(x,y)$, let $z_0=\mathcal{E}(y)$ denote the recovered target latents.
We sample Gaussian noise $\epsilon$ and a timestep $t$ from a shifted logit-normal distribution, forming
$z_t=(1-t)z_0+t\epsilon$.
The network predicts the velocity $\epsilon-z_0$, with the adapter optimized through
\begin{equation}
  \mathcal{L}
  =
  \mathbb{E}_{t,\epsilon}
  \left[
    \frac{
      \sum_i w_i
      \big\|
        v_\theta(z_t,t\mid r,c)_i
        -(\epsilon-z_0)_i
      \big\|_2^2
    }{
      \sum_i w_i
    }
  \right],
  \label{eq:loss}
\end{equation}
where the sum runs over target tokens only.
We use the masks from pair construction to set $w_i=1$ inside the latent HUD region and $w_i=0.1$ elsewhere~\cite{justdubit2026}.
This emphasizes overlay removal while retaining supervision for background preservation.
The masks weight the loss but are never supplied as network inputs; reference tokens receive no direct supervision.

\subsection{Two-stage Inference}
\label{sec:model-inference}

We use the backbone's step-distilled two-stage sampler (\cref{fig:model}, bottom). The base stage starts from Gaussian noise and performs $T_1=8$ denoising steps at reduced resolution.
A frozen latent upsampler then doubles the spatial resolution, followed by $T_2=3$ refinement steps at full resolution.

The HUD clip conditions both stages. To retain access to source detail during refinement, we re-encode the input at full resolution and attach its reference tokens again, with the same adapter active in both stages. Inference therefore uses 11 denoising steps in total, without HUD masks or externally supplied clean references.
\Cref{sec:efficiency} reports the runtime.

\section{Experiments}
\label{sec:experiments}

\subsection{Experimental Setup}
\label{sec:setup}

\noindent{\bf Benchmark.}
We evaluate GameScrub on the 100 in-the-wild clips of the Game2World benchmark~\cite{shen2026game2world}. These real gameplay recordings span distinct game titles and contain human-verified HUD tracks but no clean reference videos. The benchmark also includes 100 synthetic clips composited with the HUD asset library and renderer used for Game2World training. We focus on the in-the-wild split, which directly evaluates removal of HUDs from real footage. Supplementary material reports the synthetic split.

\noindent{\bf Evaluation protocol.}
We follow the benchmark's multimodal-judge protocol~\cite{shen2026game2world}. Given the source frame, its annotated HUD boxes, and the edited frame, the judge evaluates each annotated element. UI measures the removal rate regardless of artifacts; Clean counts artifact-free removals; and artifact-adjusted removal (AAR) assigns scores of 1, 0.5, or 0 to successful removals with no, minor, or major artifacts, respectively, and 0 to failed removals. Background preservation (BG) scores unchanged, mildly altered, and severely altered regions outside the HUD as 1, 0.5, and 0. Scores are averaged within each clip and then across clips. Using the benchmark judge (GPT-5.6 Terra), we report Overall $=(\mathrm{AAR}+\mathrm{BG})/2$ on the in-the-wild split.

\begin{table*}[t]
  \caption{{\bf HUD removal on the in-the-wild split of the Game2World benchmark} (100 real clips). UI, Clean, AAR and BG follow the protocol and judge of~\cite{shen2026game2world}; Overall is $(\mathrm{AAR}+\mathrm{BG})/2$. LPIPS and MAE are measured against the source outside the annotated HUD boxes. JEV is the mean HUD-presence probability returned by a separate decision model and YOLO the reduction in detections by the HUD detector used in GameScrub-Data (a coverage measure, see text). GameCleaner is reported in its two configurations that take no reference image at inference. $^{*}$requires a HUD mask at inference and is given the annotated boxes. $^{\dagger}$the training-free pipeline that produces GameScrub's training targets, here given the annotated boxes as its mask; it is shown for reference and excluded from the bolding. Best per column among the compared methods in bold.}
  \label{tab:main}
  \centering
  \small
  \setlength{\tabcolsep}{5pt}
  \resizebox{\textwidth}{!}{%
  \begin{tabular}{@{}lcccccccccc@{}}
    \toprule
    \textbf{Method} & \textbf{Overall$\uparrow$} & \textbf{AAR$\uparrow$} & \textbf{UI$\uparrow$} & \textbf{Clean$\uparrow$} & \textbf{BG$\uparrow$} & \textbf{LPIPS$\downarrow$} & \textbf{MAE$\downarrow$} & \textbf{JEV$\downarrow$} & \textbf{YOLO (\%)$\uparrow$} \\
    \midrule
    \multicolumn{10}{@{}l}{\emph{General video editors}} \\
    LoomVideo~\cite{wu2026loomvideo}                       & 60.58 & 45.75 & 52.38 & 43.03 & 75.40 & 0.289 & 16.72 & 0.702 & 48.3 \\
    JoyAI~\cite{joyai2026}                                 & 57.89 & 51.22 & 58.90 & 48.35 & 64.55 & 0.251 & 10.58 & 0.631 & 58.6 \\
    Lucy-Edit~\cite{decart2025lucyedit}                    & 50.33 &  5.06 &  5.92 &  4.19 & 95.60 & 0.233 & 11.68 & 0.911 &  6.2 \\
    Aurora~\cite{yu2026aurora}                             & 39.84 & 39.32 & 42.72 & 38.15 & 40.35 & 0.380 & 32.74 & 0.727 & 38.0 \\
    Kiwi-Edit~\cite{lin2026kiwiedit}                       & 37.46 & 59.96 & 66.63 & 58.05 & 14.95 & 0.453 & 27.36 & 0.522 & 61.7 \\
    OmniWeaving~\cite{pan2026omniweaving}                  & 25.52 &  8.19 & 12.47 &  6.81 & 42.85 & 0.264 & 12.63 & 0.880 & 15.1 \\
    LTX~\cite{hacohen2024ltxvideo} & 14.92 & 29.18 & 39.05 & 28.49 &  0.65 & 0.416 & 21.01 & 0.925 & $ 7.5$ \\
    VACE-1.3B~\cite{jiang2025vace}                         &  2.48 &  4.96 &  7.40 &  4.61 &  0.00 & 0.457 & 27.09 & 0.895 & 11.2 \\
    VACE-14B~\cite{jiang2025vace}                          &  1.59 &  1.17 &  1.67 &  0.97 &  2.00 & 0.287 & 18.92 & 0.943 &  2.2 \\
    \midrule
    \multicolumn{10}{@{}l}{\emph{Object removal}} \\
    EffectErase$^{*}$~\cite{fu2026effecterase}              & 49.97 & 19.39 & \textbf{91.68} & 16.99 & 80.55 & 0.289 & 16.98 & 0.542 & 65.5 \\
    \midrule
    \multicolumn{10}{@{}l}{\emph{HUD removal}} \\
    GameCleaner~\cite{shen2026game2world}                  & 73.84 & 47.98 & 49.30 & 47.38 & 99.70 & \textbf{0.077} & \textbf{3.45} & 0.701 & 70.3 \\
    GameCleaner (w.o. Ref)~\cite{shen2026game2world}        & 78.88 & 57.86 & 59.07 & 57.34 & \textbf{99.90} & 0.082 & 3.66 & 0.652 & 74.1 \\
    \textbf{GameScrub (ours)}                              & \HL\textbf{82.13} & \HL\textbf{71.95} & \HL90.06 & \HL\textbf{63.04} & \HL92.30 & \HL0.148 & \HL5.66 & \HL\textbf{0.203} & \HL\textbf{91.3} \\
    \midrule
    \multicolumn{10}{@{}l}{\emph{Data pipeline (source of GameScrub's targets, for reference)}} \\
    GameScrub-Data$^{*\dagger}$                              & 90.65 & 85.69 & 97.01 & 80.41 & 95.60 & 0.138 & 4.89 & 0.120 & 95.6 \\
    \bottomrule
  \end{tabular}
  }
\end{table*}

\noindent{\bf Auxiliary metrics.}
We additionally measure source preservation with LPIPS~\cite{zhang2018lpips} and MAE outside the annotated HUD boxes (MAE uses the 0--255 pixel range). JEV is the mean HUD-presence probability assigned to output frames by a separate decision model (GPT-6 Luna). YOLO measures the percentage reduction in HUD detections using the YOLO11n detector employed to construct GameScrub-Data. JEV and YOLO provide complementary evidence, although both remain model-based and YOLO shares the detector used in our data pipeline.


\noindent{\bf Baselines.}
We compare against the general video editors Aurora~\cite{yu2026aurora}, Kiwi-Edit~\cite{lin2026kiwiedit}, LoomVideo~\cite{wu2026loomvideo}, Lucy-Edit~\cite{decart2025lucyedit}, OmniWeaving~\cite{pan2026omniweaving}, LTX~\cite{hacohen2024ltxvideo}, VACE~\cite{jiang2025vace} (1.3B and 14B), and JoyAI~\cite{joyai2026}. We also evaluate EffectErase~\cite{fu2026effecterase}, an object-removal method, and GameCleaner~\cite{shen2026game2world}, the prior task-specific model. GameCleaner is tested in its standard configuration and its reference-free inference variant (w.o. Ref); the configuration requiring a clean reference image is excluded. All general editors use text-only instructions without reference images or HUD masks. EffectErase is given the benchmark's annotated boxes and is therefore not a mask-free comparison. Following~\cite{shen2026game2world}, baseline outputs contain 81 frames at 16\,fps and $1280\times720$ resolution; all outputs are sampled at 2\,fps and resized to $1280\times720$ for evaluation. Supplementary material lists all baseline settings and metric definitions.

\begin{figure}[t]
  \centering
  \includegraphics[width=0.7\linewidth]{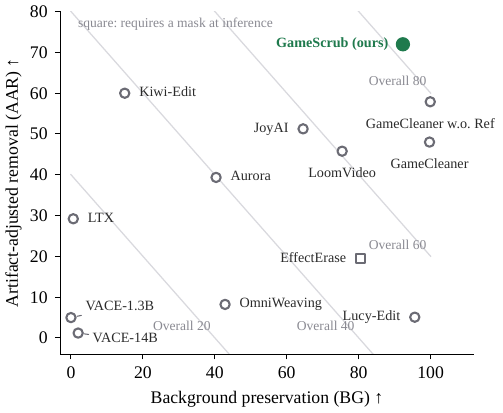}
  \caption{{\bf Removal against preservation on the in-the-wild split.} Diagonals are lines of constant Overall. Methods that preserve the background best are those that remove the least; GameScrub is the only mask-free method above 70 AAR and above 90 BG.}
  \label{fig:tradeoff}
\end{figure}

\noindent{\bf Implementation.}
GameScrub builds on LTX-2.5~\cite{ltx25,hacohen2026ltx2}, a step-distilled video diffusion transformer with two-stage sampling. We train a rank-128 in-context LoRA while freezing the backbone and autoencoder. LTX-2.5 supports variable clip lengths and substantially fewer sampling steps than the 50-step backbone used by GameCleaner. Training uses approximately 8K pairs from GameScrub-Data, AdamW with an initial learning rate of $2\times10^{-4}$ and linear decay, and a global batch size of 32 across eight H200 GPUs for 300 steps. Inference uses 8 steps at $640\times352$ followed by 3 refinement steps at $1280\times704$, with the source clip re-encoded at full resolution and the adapter active in both stages (\cref{sec:model-inference}).


\subsection{Comparison with Prior Work}
\label{sec:results-main}

\Cref{tab:main} summarizes quantitative results, and \cref{fig:tradeoff} visualizes the trade-off between HUD removal and background preservation.

\begin{figure*}[t]
  \centering
  \includegraphics[width=\linewidth]{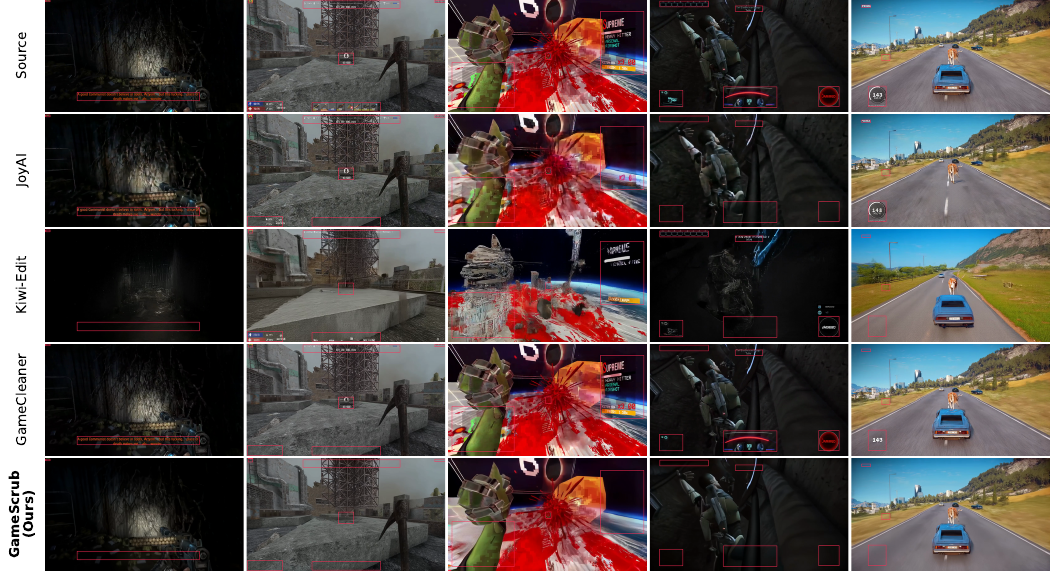}
  \caption{{\bf Qualitative comparison} on in-the-wild benchmark clips. Red rectangles mark the annotated HUD regions, drawn for reference.}
  \label{fig:qual}
\end{figure*}

\noindent{\bf HUD removal.}
GameScrub's main advantage is more complete HUD removal (\cref{fig:qual}): AAR reaches 71.95, compared with 57.86 for GameCleaner (w.o. Ref) and 47.98 for standard GameCleaner (paired statistics in supplementary). It also has the highest Overall among mask-free methods (82.13, against 78.88 for GameCleaner (w.o. Ref) and 60.58 for LoomVideo, the strongest general editor). GameScrub removes 90.06\% of annotated elements, versus 59.07\% for GameCleaner (w.o. Ref), while also improving artifact-free removal (63.04 versus 57.34). EffectErase attains a slightly higher raw removal rate (91.68), but requires a HUD mask and leaves substantial artifacts, reducing its AAR to 19.39. On the 30 benchmark clips whose game is identified with confidence and matches no title or franchise in our training footage, GameScrub reaches 72.69 AAR against 57.27 for GameCleaner (w.o. Ref) (see supplementary).

\noindent{\bf Background preservation.}
Greater removal comes at a cost to source fidelity. GameScrub scores 92.30 BG, below GameCleaner (99.70) and GameCleaner (w.o. Ref) (99.90); its LPIPS and MAE are also higher than those of GameCleaner. GameScrub regenerates the full frame, which introduces changes outside HUD regions, and decodes at a height of 704 pixels, which can reduce fine detail. Conversely, methods with near-perfect BG leave many HUD elements untouched, whereas Kiwi-Edit removes more at the expense of severe background changes. As shown in \cref{fig:tradeoff}, GameScrub is the only mask-free method with AAR above 70 and BG above 90.


\noindent{\bf Auxiliary evidence.}
The additional metrics support the removal results. GameScrub reduces HUD-presence probability to 0.203, compared with 0.652 and 0.701 for the two GameCleaner variants. It also removes 91.3\% of the HUD detections identified by YOLO, versus 74.1\% and 70.3\% for GameCleaner. These measures corroborate the main trend.


\noindent{\bf Quality of recovered training targets.}
\label{sec:results-teacher}
We evaluate GameScrub-Data itself using the benchmark's annotated HUD boxes to isolate target recovery from mask estimation. This mask-conditioned pipeline achieves 97.01 UI, 80.41 Clean, 85.69 AAR, and 95.60 BG (\cref{tab:main}). It is an upper reference for the target-construction procedure, not a competing mask-free method. Without masks, GameScrub retains approximately 93\% of the pipeline's removal rate and 84\% of its AAR. The remaining gap partly reflects the imperfect recovered targets: even with annotated masks, approximately 17\% of successful removals contain visible artifacts, compared with approximately 30\% for GameScrub.


\noindent{\bf Efficiency.}
\label{sec:efficiency}
GameScrub uses 11 sampling steps per clip, compared with 50 for GameCleaner. On a single H200 GPU, processing a 5-second 720p clip takes approximately 27\,s versus 57\,s, a $2.1\times$ speed-up. This corresponds to 5.4 versus 11.4 GPU-hours per hour of footage.

\subsection{GameScrub Ablations}
\label{sec:ablations-model}

\noindent{\bf Second-stage conditioning.}
We ablate the refinement stage while keeping the trained adapter and first-stage output fixed. Stopping after the low-resolution stage yields 0.253 LPIPS and 7.25 MAE outside the HUD. Running the second stage without the source reference or adapter further degrades preservation to 0.428 LPIPS and 19.06 MAE: refinement then regenerates background detail without access to the original video. Reintroducing the full-resolution source reference and the trained adapter improves these metrics to 0.148 and 5.66, respectively (\cref{tab:abl-stage2}). This result shows that the second stage improves fidelity only when equipped with source conditioning and the adapted model.


\begin{table}
  \small
  \setlength{\tabcolsep}{6pt}
  \caption{{\bf Second-stage configuration.} LPIPS and MAE against the source outside the annotated HUD boxes on the in-the-wild split; lower is better. The last row is the model reported in \cref{tab:main}.}
  \label{tab:abl-stage2}
  \centering
  \begin{tabular}{@{}lcc@{}}
    \toprule
    \textbf{Inference configuration} & \textbf{LPIPS$\downarrow$} & \textbf{MAE$\downarrow$} \\
    \midrule
    Stage 1 only & 0.253 & 7.25 \\
    + stage 2, plain (no reference, no adapter) & 0.428 & 19.06 \\
    + stage 2 with reference and adapter (ours) & \textbf{0.148} & \textbf{5.66} \\
    \bottomrule
  \end{tabular}
\end{table}


\section{Conclusion}
\label{sec:conclusion}

We introduce GameScrub, a mask-free HUD remover for real gameplay videos. Rather than synthesizing HUD overlays on clean footage, GameScrub-Data recovers approximate clean targets from real recordings using flow-guided pre-filling and frozen video diffusion. Trained on these pairs, GameScrub removes HUDs without masks or reference images in 11 denoising steps. On the in-the-wild Game2World benchmark, it improves artifact-adjusted removal from 57.86 to 71.95 over the strongest reference-free GameCleaner variant, while achieving a 2.1$\times$ inference speedup. These results show that real inputs with recovered targets are sufficient to learn HUD removal on real footage, and that the remaining limit is the quality of the recovered targets rather than the recognition of real HUDs.

\noindent{\bf Limitations.}
GameScrub's mask-free design reveals a trade-off between HUD removal and background preservation: full-frame generation improves removal coverage but may alter scene details that should remain unchanged. This motivates spatially selective generation that localizes edits without requiring explicit masks. Moreover, recovered training targets become less reliable when HUDs escape detection or obscure content with insufficient visual evidence. Uncertainty-aware reconstruction and supervision offer a promising direction for improving target quality. Finally, while GameScrub enables large-scale cleaning of real gameplay footage, its impact on downstream world model training remains unexplored. Investigating how HUD removal affects the learning of scene dynamics and interactive behavior is an important next step.

\clearpage
\appendix



\section{Additional Results}\label{sec:appendix-results}


\noindent This supplement gives the settings of GameScrub-Data (\cref{sec:supp-data}), ablates its steps (\cref{sec:supp-ablation}), lists the settings of the baselines (\cref{sec:supp-baselines}), specifies every metric (\cref{sec:supp-eval}), and reports per-clip statistics (\cref{sec:supp-stats}), results on the synthetic split (\cref{sec:supp-synthetic}), a control that retrains GameCleaner on our pairs (\cref{sec:supp-control}), the overlap between training and benchmark games (\cref{sec:supp-overlap}), and qualitative comparisons by HUD type (\cref{sec:supp-qual}).

\subsection{GameScrub-Data: settings}
\label{sec:supp-data}

The dataset is not released; this section gives the settings of the pipeline that produced it.

\paragraph{Source footage.}
The training set holds 8{,}522 windows of 145 frames at 30 fps (4.83\,s), resized to $640{\times}352$ with area interpolation.
3{,}599 windows come from 439 publicly available gameplay videos and 4{,}923 from 1{,}439 gameplay recordings made in-house.
A game title is known for 7{,}737 of the 8{,}522 windows (91\%), covering about 300 distinct titles, 219 of them from the public videos.
Source clips of 60\,s are cut into non-overlapping windows, and windows in which the detector finds no HUD are not used.

\paragraph{HUD detector.}
The detector is YOLO11n fine-tuned for a single class, removable HUD overlay, at $736{\times}1312$ for 100 epochs with batch size 16; we use the checkpoint with the best validation score (epoch 52).
Its training set is 14{,}632 frames labelled automatically by a multimodal language model, split into 14{,}532 for training and 100 for validation.
On the validation frames it reaches a precision of 0.788, a recall of 0.766, an mAP$_{50}$ of 0.827 and an mAP$_{50\text{-}95}$ of 0.610; these figures are measured against the automatic labels, not against human annotation.
Detections with confidence of at least 0.25 are kept.

\paragraph{From boxes to masks.}
Per-frame detections are first consolidated over time: boxes are united across neighbouring frames and isolated outliers are repaired.
The boxes are then rasterised as filled rectangles with no spatial dilation, and the mask is dilated in time only: a pixel is masked in a frame if a box covers it in any frame within 16 frames before or after, which closes the gaps left by a few missed detections.
A latent cell covers $32{\times}32$ pixels and belongs to the latent mask $M$ if more than 2\% of its area is masked.

\paragraph{Caption.}
The captioner is gpt-6-luna at medium reasoning effort.
It receives 8 frames sampled evenly across the clip, each with the masked region painted solid black, so it never sees the HUD.
Its system prompt is reproduced below, with $n$ the number of frames and $d$ the clip duration.
A second note tells the model that the marked regions will be repainted, that it must not describe or hint at their current content or mention a box or a solid patch, and that it should describe them as the natural continuation of the surrounding scene.
\begin{quote}
\small
You are shown $n$ frames sampled in chronological order from a $d$-second single-take video clip. Write ONE caption describing THIS clip, in the EXACT style of the training captions used for the LTX-2 video model. The caption will guide a detail-preserving generation of the same clip, so it must be faithful: describe only what is actually visible in the frames and how it actually changes across them. Never invent events, objects, text or camera moves that are not visible.

Match this captioning style precisely:

1. Begin immediately with the action or visual detail. Do NOT use ``The scene opens\ldots'', ``We see\ldots'', ``There is\ldots'', ``The video shows\ldots''.

2. Objective, observable description only. Do not infer emotions or intentions.

3. Full visual detail: environment, character appearance, spatial positioning of all elements.

4. Precise motion and cinematic description woven into the prose: shot type, camera motion, camera viewpoint.

5. Strict chronological, real-time flow using transitions like ``Initially\ldots'', ``A moment later\ldots'', ``By the end\ldots''.

6. Visual medium and condition, briefly, near the end (digital video, animation, etc.\ --- whatever is actually visible).

7. One single continuous paragraph, 150--220 words, never exceeding 300 words.

Audio: [a fixed sentence telling the model not to describe or invent sound].

Return JSON:\\ \texttt{\{"prompt": "<the caption>"\}}
\end{quote}

\paragraph{Pre-fill.}
The pre-fill is ProPainter~\cite{zhou2023propainter}, which erases the HUD from the $640{\times}352$ clip given the pixel mask.
We run it with its default settings: mask dilation 4, reference stride 10, neighbour length 10, sub-video length 80, 20 RAFT iterations, full precision and no resizing.
ProPainter dilates the mask internally, so the fill extends slightly beyond the mask; the composite below uses the undilated mask.

\paragraph{Masked latent diffusion.}
The backbone is the distilled LTX-2.5 transformer (22B parameters, bf16) with no adapter.
Sampling takes 8 steps through the noise levels 1, 0.99375, 0.9875, 0.98125, 0.975, 0.909375, 0.725, 0.421875 and 0, with one transformer call per step, no classifier-free guidance and no negative prompt.
Regeneration is binary per latent cell: the cells in $M$ start from pure noise and are denoised, and every other cell is held at the encoding of the pre-filled clip (\cref{eq:snap}).
The seed is fixed.

\paragraph{Composite.}
The target is $y = M_{\mathrm{px}} \odot \hat{x} + (1 - M_{\mathrm{px}}) \odot x$ with a hard cut and no feathering (\cref{sec:data-diffusion}).

\paragraph{On the benchmark.}
For the last row of \cref{tab:main} and for \cref{tab:supp-ablation} the same pipeline runs on the benchmark clips with the annotated boxes in place of the detector boxes.
Its $640{\times}352$ output is upscaled to the scoring resolution.

\subsection{Ablation of GameScrub-Data}
\label{sec:supp-ablation}

\begin{table*}[t]
  \caption{Ablation of GameScrub-Data on the in-the-wild split (100 clips), with the annotated boxes as mask. \emph{w/o pre-fill} encodes the raw clip in place of the ProPainter output. \emph{w/o diffusion} takes the ProPainter output as the result. \emph{Generic prompt} replaces the per-clip caption with one fixed instruction. \emph{w/o composite} uses the decoded clip as the result instead of pasting its HUD region into the source. Metrics as in \cref{tab:main}.}
  \label{tab:supp-ablation}
  \centering
  \small
  \begin{tabular}{@{}lccccccc@{}}
    \toprule
    \textbf{Variant} & \textbf{Overall$\uparrow$} & \textbf{AAR$\uparrow$} & \textbf{UI$\uparrow$} & \textbf{Clean$\uparrow$} & \textbf{BG$\uparrow$} & \textbf{JEV$\downarrow$} & \textbf{YOLO (\%)$\uparrow$} \\
    \midrule
    Full pipeline   & 90.65 & 85.69 & 97.01 & 80.41 & 95.60 & 0.120 & 95.6 \\
    \midrule
    w/o pre-fill    & 76.79 & 59.23 & 70.06 & 55.87 & 94.35 & 0.801 & 38.0 \\
    w/o diffusion   & 79.96 & 61.01 & 97.11 & 48.16 & 98.90 & 0.117 & 95.6 \\
    Generic prompt  & 88.50 & 81.85 & 95.73 & 74.72 & 95.15 & 0.134 & 95.2 \\
    w/o composite   & 86.67 & 86.20 & 93.98 & 81.83 & 87.15 & 0.070 & 97.0 \\
    \bottomrule
  \end{tabular}
\end{table*}

\Cref{tab:supp-ablation} removes or replaces one step of the pipeline at a time, under the protocol of the last row of \cref{tab:main}.

\paragraph{Pre-fill.}
Without the pre-fill the pipeline removes 70.06\% of the annotated elements instead of 97.01\%, and AAR falls from 85.69 to 59.23.
The two measures that do not use the judge move the same way: the reduction in detections falls from 95.6\% to 38.0\%, and the HUD-presence probability rises from 0.120 to 0.801.
BG is almost unchanged (94.35 against 95.60), because the composite still copies the source outside the mask.
This is the failure shown on one clip in \cref{fig:leak}: with the raw clip as conditioning the HUD survives in the output, and the pre-fill is what prevents it.

\paragraph{Diffusion.}
The pre-fill alone is not a substitute for the pipeline.
Taking the ProPainter output as the result removes as many elements (97.11\% UI) but only 48.16\% without an artifact, against 80.41\% for the full pipeline, and AAR is 61.01 against 85.69.
The two steps are therefore complementary: the pre-fill decides whether the HUD is removed, and the diffusion step decides whether the removal is clean.
The diffusion step costs some preservation, 95.60 BG against 98.90.
JEV and the detector measure do not separate this variant from the full pipeline (0.117 and 95.6\% against 0.120 and 95.6\%): they register whether a HUD is still present, not the quality of what replaces it, which only Clean and AAR capture.

\paragraph{Caption.}
Replacing the per-clip caption with the fixed instruction ``Remove the game HUD overlay from this footage, revealing the clean underlying gameplay video.'' lowers Clean from 80.41 to 74.72 and AAR from 85.69 to 81.85, while UI and BG change little (95.73 and 95.15).
The per-clip caption thus improves what is generated in the HUD region.
This variant replaces the caption altogether; it does not isolate the effect of hiding the HUD from the captioner.

\paragraph{Composite.}
Without the composite the removal measures change little, and four of the five are slightly higher (AAR 86.20 against 85.69, Clean 81.83 against 80.41, and both supporting measures), while UI is lower (93.98 against 97.01).
Preservation is what the composite buys: BG rises from 87.15 to 95.60, and Overall from 86.67 to 90.65.
For training it has a second role: the target equals the source outside the mask, so GameScrub is never supervised to change the background (\cref{sec:data-diffusion}).

\subsection{Baselines}
\label{sec:supp-baselines}

\begin{table*}[t]
  \caption{Settings of the baselines. All receive the same instruction and the same input clip (81 frames at 16 fps, $1280{\times}720$), generate at their native resolution with the settings listed, and are rescaled to $1280{\times}720$ for scoring. $^{\ddagger}$FlowEdit~\cite{kulikov2024flowedit} with source guidance 3 and target guidance 6.5.}
  \label{tab:supp-baselines}
  \centering
  \small
  \resizebox{\textwidth}{!}{%
  \begin{tabular}{@{}llcccc@{}}
    \toprule
    \textbf{Method} & \textbf{Checkpoint} & \textbf{Native output} & \textbf{Steps} & \textbf{Guidance} & \textbf{Mask} \\
    \midrule
    Aurora~\cite{yu2026aurora}              & Aurora editor, Wan2.2-TI2V-5B, Qwen3.5-4B          & $1280{\times}704$ & 50 & 2.0 & no \\
    JoyAI~\cite{joyai2026}                  & \texttt{joyai\_video\_edit\_dit\_0811} (distilled) & $1248{\times}720$ & 2  & 1.0 & no \\
    Kiwi-Edit~\cite{lin2026kiwiedit}        & \texttt{kiwi-edit-5b-instruct-reference}           & $1280{\times}720$ & 50 & 5.0 & no \\
    LoomVideo~\cite{wu2026loomvideo}        & LoomVideo (RL variant), Wan2.2-TI2V-5B, Qwen3-VL-8B & $832{\times}480$  & 50 & default & no \\
    LTX                                     & LTX-Video 2B v0.9.5                                & $1024{\times}576$ & 30 & 3 / 6.5$^{\ddagger}$ & no \\
    OmniWeaving~\cite{pan2026omniweaving}   & HY-OmniWeaving                                     & $848{\times}480$  & 50 & default & no \\
    VACE-1.3B~\cite{jiang2025vace}          & Wan2.1-VACE-1.3B                                   & $832{\times}480$  & 30 & 5.0 & no \\
    VACE-14B~\cite{jiang2025vace}           & Wan2.1-VACE-14B                                    & $1280{\times}720$ & 30 & 5.0 & no \\
    Lucy-Edit~\cite{decart2025lucyedit}     & Lucy-Edit 1.1 Dev                                  & $1280{\times}704$ & 50 & 5.0 & no \\
    EffectErase~\cite{fu2026effecterase}    & EffectErase, Wan2.1-Fun-1.3B-InP                   & $832{\times}480$  & 50 & 1.0 & yes \\
    \bottomrule
  \end{tabular}
  }
\end{table*}

\paragraph{Settings.}
Every baseline of \cref{tab:supp-baselines} receives the instruction ``Remove the gameplay HUD from the video.'', the seed 2025, and the benchmark clip conformed to 81 frames at 16 fps and $1280{\times}720$.
Each model then runs with the settings of \cref{tab:supp-baselines}, which are fixed for all clips, and its output is rescaled to $1280{\times}720$ for scoring.
All general editors, including the two VACE models, are run from the instruction alone and receive no mask, which is the setting GameScrub is used in.
EffectErase cannot run without a mask and receives the annotated boxes.
JoyAI generates 121 frames at 24 fps, which are resampled in time to 81 frames at 16 fps.
Four models generate at about 480p and one at 576p, so their LPIPS and MAE include the loss from upscaling.
The LTX row is run with FlowEdit~\cite{kulikov2024flowedit}, which takes a source and a target prompt in place of the instruction: ``A gameplay video with HUD overlays, menus, text and other on-screen interface elements on top of the game world.'' and ``The same gameplay video showing only the game world, with no HUD, menus, text or on-screen interface elements.''

\subsection{Evaluation protocol}
\label{sec:supp-eval}

\paragraph{Judge metrics.}
UI, Clean, AAR and BG are computed with the evaluation code and judge prompt of~\cite{shen2026game2world}.
The judge is gpt-5.6-terra at medium reasoning effort, called once per clip; there is no vote and no averaging over repeated calls.
Frames are sampled at 2 per second from the start of the source clip, and each is paired with the output frame nearest in time.
For every sampled frame the judge sees the source and output frames one above the other, with the annotated HUD boxes drawn in red on this evidence image only, and, for each annotated element, a crop of the source and of the output around its box with 1.5 times the box as context, plus an enlarged view when the box is at most 96 pixels.
It labels each element as \emph{removed}, \emph{present} or \emph{uncertain}, with an artifact grade of \emph{none}, \emph{minor} or \emph{major}, and each frame as \emph{preserved}, \emph{minor change}, \emph{major change} or \emph{invalid}.
With $r_e$ the removal label and $a_e$ the artifact grade of element $e$, and $p_f$ the label of frame $f$,
\begin{align}
  \mathrm{UI}    &= \operatorname{mean}_e\, \mathbf{1}[r_e = \text{removed}], \\
  \mathrm{Clean} &= \operatorname{mean}_e\, \mathbf{1}[r_e = \text{removed},\ a_e = \text{none}], \\
  \mathrm{AAR}   &= \operatorname{mean}_e\, \mathbf{1}[r_e = \text{removed}]\; w(a_e), \\
  \mathrm{BG}    &= \operatorname{mean}_f\, v(p_f),
\end{align}
with $w(\text{none}) = 1$, $w(\text{minor}) = 0.5$ and $w = 0$ otherwise, and $v(\text{preserved}) = 1$, $v(\text{minor change}) = 0.5$ and $v = 0$ otherwise.
Scores are averaged over the elements of a frame, then over the frames of a clip, then over clips, and reported in percent.

\paragraph{JEV.}
JEV asks a decision model, gpt-6-luna, one yes/no question about each sampled frame of the output clip and reads the probability of ``yes''.
The model sees only the output frame: no source frame, no annotation and no box.
The probabilities are averaged over the frames of a clip without thresholding, and then over clips.
The question is:
\begin{quote}
\small
Does this game image have any HUD, UI, or overlay elements on it? Only count 2D screen-space graphics layered on top of the gameplay view -- things like health/status bars, minimaps or radars, crosshairs, ammo/ability counters, notifications or popups, chat or subtitles, scoreboards, menu controls, watermarks, or streamer/donation overlays. Do NOT count in-world diegetic screens, monitors, or displays that are rendered as part of the 3D game environment itself (e.g.\ a cockpit computer screen, an in-game terminal, a TV inside the game world) -- these are part of the scene, not an overlay, even if they show text or graphics. If this frame is a loading screen, cutscene, or menu transition rather than actual gameplay, judge only whether gameplay-overlay HUD is present, not any informational text that's part of the loading/menu screen itself.
\end{quote}

\paragraph{Detector measure (YOLO).}
The detector of \cref{sec:supp-data} is run on sampled frames of the source clip and of the output clip, each resized to a width of 1312 pixels, and detections with confidence of at least 0.25 are counted.
With $N_{\mathrm{in}}$ and $N_{\mathrm{out}}$ the total counts over the sampled frames of a clip, the measure is $100\,(N_{\mathrm{in}} - N_{\mathrm{out}})/N_{\mathrm{in}}$; a clip with $N_{\mathrm{in}} = 0$ is left out.
Because the same detector supplies the masks of GameScrub-Data, this measure is not independent of GameScrub's training data, and we use it only as a coverage measure.

\paragraph{LPIPS and MAE.}
Both compare the output with the source on background pixels, that is, outside the annotated HUD boxes, after the two are brought to the same resolution.
LPIPS~\cite{zhang2018lpips} uses the AlexNet backbone and is evaluated as a spatial map, which is averaged over the background pixels; MAE is the mean absolute difference on the 0--255 scale.
Values are averaged over the frames of a clip and then over clips.

\subsection{Per-clip statistics}
\label{sec:supp-stats}

\begin{table}[t]
  \caption{Paired per-clip comparison of GameScrub with the two GameCleaner rows of \cref{tab:main}, 100 clips per split. $\Delta$ is GameScrub minus GameCleaner; the interval is a 95\% percentile bootstrap interval of the mean per-clip difference (20{,}000 resamples of the clips). W/L/T counts the clips on which GameScrub scores higher, lower, or the same.}
  \label{tab:supp-stats}
  \centering
  \footnotesize
  \setlength{\tabcolsep}{2.2pt}
  \begin{tabular}{@{}lccccc@{}}
    \toprule
    & \textbf{GameCleaner} & \textbf{GameScrub} & \textbf{$\Delta$} & \textbf{95\% interval} & \textbf{W/L/T} \\
    \midrule
    \multicolumn{6}{@{}l}{\emph{In-the-wild, against GameCleaner}} \\
    Overall & 73.84 & 82.13 & $+8.29$  & $[+3.8,\,+12.6]$  & 64/21/15 \\
    AAR     & 47.98 & 71.95 & $+23.97$ & $[+16.8,\,+31.0]$ & 67/14/19 \\
    BG      & 99.70 & 92.30 & $-7.40$  & $[-12.2,\,-3.3]$  & 3/15/82 \\
    \midrule
    \multicolumn{6}{@{}l}{\emph{In-the-wild, against GameCleaner (w.o. Ref)}} \\
    Overall & 78.88 & 82.13 & $+3.25$  & $[-1.5,\,+7.9]$   & 48/32/20 \\
    AAR     & 57.86 & 71.95 & $+14.09$ & $[+6.3,\,+21.8]$  & 51/28/21 \\
    BG      & 99.90 & 92.30 & $-7.60$  & $[-12.4,\,-3.5]$  & 1/15/84 \\
    \midrule
    \multicolumn{6}{@{}l}{\emph{Synthetic, against GameCleaner}} \\
    Overall & 97.26 & 90.28 & $-6.98$  & $[-9.7,\,-4.2]$   & 6/57/37 \\
    AAR     & 95.51 & 82.21 & $-13.30$ & $[-17.5,\,-9.0]$  & 5/57/38 \\
    BG      & 99.00 & 98.35 & $-0.65$  & $[-3.7,\,+2.4]$   & 1/3/96 \\
    \midrule
    \multicolumn{6}{@{}l}{\emph{Synthetic, against GameCleaner (w.o. Ref)}} \\
    Overall & 97.08 & 90.28 & $-6.80$  & $[-9.5,\,-4.0]$   & 7/60/33 \\
    AAR     & 95.66 & 82.21 & $-13.45$ & $[-17.3,\,-9.7]$  & 6/59/35 \\
    BG      & 98.50 & 98.35 & $-0.15$  & $[-3.3,\,+3.0]$   & 2/3/95 \\
    \bottomrule
  \end{tabular}
\end{table}

\Cref{tab:supp-stats} compares GameScrub with each GameCleaner row of \cref{tab:main} clip by clip.
On the in-the-wild split and against the standard GameCleaner, GameScrub leads Overall by 8.29 points, with a 95\% interval of 3.8 to 12.6, and AAR by 23.97 points (16.8 to 31.0), and trails in BG by 7.40 points (3.3 to 12.2).
Against GameCleaner (w.o. Ref), GameScrub leads AAR by 14.09 points (6.3 to 21.8) and the raw removal rate UI by 30.99 points (23.9 to 38.2), and trails in BG by 7.60 points (3.5 to 12.4).
The two effects partly offset in Overall: the difference of 3.25 points has an interval of $-1.5$ to 7.9, which includes zero.
The advantage of GameScrub over GameCleaner (w.o. Ref) is therefore in removal, and on Overall the two are not separated by a benchmark of 100 clips.

\subsection{Synthetic split}
\label{sec:supp-synthetic}

The main paper reports the in-the-wild split, because the task is real footage.
The lower half of \cref{tab:supp-stats} gives the synthetic split of the same benchmark.
GameCleaner leads there in both configurations: 97.26 and 97.08 Overall against 90.28 for GameScrub, and 95.51 and 95.66 AAR against 82.21, with intervals that exclude zero.
The differences in BG are not distinguishable from zero (99.00 and 98.50 against 98.35).
The synthetic split is composited with the HUD asset library and renderer that generate the pairs GameCleaner is trained on~\cite{shen2026game2world}, so it tests GameCleaner on HUDs from its training distribution and GameScrub, which was trained on real HUDs only, outside its own.
The two splits together show how each model transfers: from the synthetic split to the in-the-wild split GameCleaner loses 47.53 AAR points (95.51 to 47.98) and GameCleaner (w.o. Ref) 37.80 (95.66 to 57.86), while GameScrub loses 10.26 (82.21 to 71.95).

\subsection{Same architecture, different training data}
\label{sec:supp-control}

\begin{table*}[t]
  \caption{GameCleaner retrained on GameScrub-Data pairs, on the in-the-wild split; columns as in \cref{tab:main}. The first, second and last rows are those of \cref{tab:main}. The third row keeps the architecture and training configuration of GameCleaner and changes only the training pairs.}
  \label{tab:supp-control}
  \centering
  \small
  \setlength{\tabcolsep}{4.5pt}
  \resizebox{\textwidth}{!}{%
  \begin{tabular}{@{}llccccccccc@{}}
    \toprule
    \textbf{Model} & \textbf{Training pairs} & \textbf{Overall$\uparrow$} & \textbf{AAR$\uparrow$} & \textbf{UI$\uparrow$} & \textbf{Clean$\uparrow$} & \textbf{BG$\uparrow$} & \textbf{LPIPS$\downarrow$} & \textbf{MAE$\downarrow$} & \textbf{JEV$\downarrow$} & \textbf{YOLO (\%)$\uparrow$} \\
    \midrule
    GameCleaner              & synthetic      & 73.84 & 47.98 & 49.30 & 47.38 & 99.70 & 0.077 & 3.45 & 0.701 & 70.3 \\
    GameCleaner (w.o. Ref)   & synthetic      & 78.88 & 57.86 & 59.07 & 57.34 & 99.90 & 0.082 & 3.66 & 0.652 & 74.1 \\
    GameCleaner (our pairs)  & GameScrub-Data & 76.80 & 69.55 & 87.50 & 63.80 & 84.05 & 0.240 & 6.06 & 0.256 & 94.7 \\
    \midrule
    GameScrub                & GameScrub-Data & 82.13 & 71.95 & 90.06 & 63.04 & 92.30 & 0.148 & 5.66 & 0.203 & 91.3 \\
    \bottomrule
  \end{tabular}
  }
\end{table*}

GameScrub and GameCleaner differ in both their training pairs and their backbone, so \cref{tab:main} does not show which of the two the difference in removal comes from.
We therefore retrain GameCleaner on our pairs.
The architecture, the initialization and the training configuration are those reported for the standard GameCleaner~\cite{shen2026game2world}: a rank-64 LoRA on the DiT with the multimodal encoder and connector frozen, a global batch size of 32, a learning rate of $10^{-4}$ and 3{,}000 steps.
Only the training pairs change, from the synthetic pairs of~\cite{shen2026game2world} to the GameScrub-Data pairs that train GameScrub (\cref{sec:supp-data}).
The retrained model is run and scored as the GameCleaner rows of \cref{tab:main}.

\paragraph{Removal follows the training pairs.}
With the architecture fixed, replacing synthetic pairs by recovered pairs raises AAR from 47.98 to 69.55 and UI from 49.30 to 87.50, and lowers JEV from 0.701 to 0.256 (\cref{tab:supp-control}).
The retrained model is also 11.69 AAR points above GameCleaner (w.o. Ref).
GameScrub is 23.97 AAR points above the standard GameCleaner, and the retrained model recovers 21.57 of them on GameCleaner's own architecture.

\paragraph{So does the cost in preservation.}
The same change lowers BG from 99.70 to 84.05 and raises LPIPS from 0.077 to 0.240.
Overall therefore rises only from 73.84 to 76.80, which is below GameCleaner (w.o. Ref) at 78.88.

\paragraph{The two models trained on the same pairs.}
GameScrub and the retrained GameCleaner are close in removal: 71.95 and 69.55 AAR, and 63.04 and 63.80 Clean.
They differ in preservation, with 92.30 against 84.05 BG and 0.148 against 0.240 LPIPS, and in cost, with 11 sampling steps against 50.

\paragraph{Limits.}
This experiment changes the training pairs under GameCleaner's architecture.
We do not train GameScrub's backbone on the synthetic pairs, so it shows that recovered pairs are sufficient for the gain in removal on that architecture and does not measure what synthetic pairs would give on ours.

\subsection{Overlap between training and benchmark games}
\label{sec:supp-overlap}

\begin{table*}[t]
  \caption{In-the-wild results by overlap with the training footage. Rows (i) to (iii) are cumulative: (i) the clip's game title appears in training; (ii) also the same game under another edition, expansion or compilation name; (iii) also another game of the same franchise. \emph{No match} is the rest, and the last row keeps those of them whose title is identified with confidence of at least 0.8. $\Delta$AAR is GameScrub minus GameCleaner (w.o. Ref) with a 95\% paired bootstrap interval.}
  \label{tab:supp-overlap}
  \centering
  \small
  \setlength{\tabcolsep}{4.5pt}
  \resizebox{\textwidth}{!}{%
  \begin{tabular}{@{}lccccccccccc@{}}
    \toprule
    & & \multicolumn{4}{c}{\textbf{GameScrub}} & \multicolumn{4}{c}{\textbf{GameCleaner (w.o. Ref)}} & & \\
    \cmidrule(lr){3-6}\cmidrule(lr){7-10}
    \textbf{Clips} & \textbf{$n$} & \textbf{Overall} & \textbf{AAR} & \textbf{UI} & \textbf{BG} & \textbf{Overall} & \textbf{AAR} & \textbf{UI} & \textbf{BG} & \textbf{$\Delta$AAR} & \textbf{95\% interval} \\
    \midrule
    All                                   & 100 & 82.13 & 71.95 & 90.06 & 92.30 & 78.88 & 57.86 & 59.07 & 99.90  & $+14.09$ & $[+6.3,\,+21.8]$ \\
    \midrule
    (i) same title                        & 16  & 74.05 & 54.36 & 84.92 & 93.75 & 73.97 & 48.57 & 50.29 & 99.38  & $+5.79$  & $[-15.0,\,+25.8]$ \\
    (ii) same game                        & 22  & 76.00 & 58.36 & 82.68 & 93.64 & 74.37 & 49.19 & 50.44 & 99.55  & $+9.17$  & $[-7.3,\,+24.9]$ \\
    (iii) same game or franchise          & 56  & 81.19 & 67.37 & 87.44 & 95.00 & 78.38 & 56.93 & 58.71 & 99.82  & $+10.44$ & $[-0.2,\,+21.0]$ \\
    \midrule
    No match                              & 44  & 83.33 & 77.79 & 93.40 & 88.86 & 79.52 & 59.03 & 59.51 & 100.00 & $+18.76$ & $[+8.2,\,+29.4]$ \\
    No match, title identified            & 30  & 81.51 & 72.69 & 91.73 & 90.33 & 78.64 & 57.27 & 57.97 & 100.00 & $+15.42$ & $[+2.4,\,+28.0]$ \\
    \bottomrule
  \end{tabular}
  }
\end{table*}

The benchmark does not state which game each clip shows, so we identify the games and compare them with the titles in our training footage.

\paragraph{Identifying the games.}
For each of the 100 in-the-wild clips, four frames of the source video, taken at 0.625, 1.875, 3.125 and 4.375\,s with the HUD on screen, are given together to gpt-6-luna.
The model returns the commercial title, a confidence, alternative titles and the visible evidence, and is instructed to answer ``unknown'' rather than guess.
It names a title for 98 clips, 83 of them with confidence of at least 0.8, and 92 distinct titles in total.

\paragraph{Matching.}
Titles are compared after removing case, punctuation and spacing and writing roman numerals as digits.
We match at three levels.
(i) \emph{Same title}: the clip's title appears among the training titles (16 clips).
(ii) \emph{Same game}: in addition, the same game under another edition, expansion or compilation name, for example \emph{Alan Wake} and its remaster (6 further clips).
(iii) \emph{Same franchise}: in addition, a different game of a franchise present in training, for example another \emph{Assassin's Creed}, \emph{Halo} or \emph{Uncharted} entry (34 further clips).
The remaining 44 clips match at no level; 30 of them have a title identified with confidence of at least 0.8, and 14 are unidentified or identified with low confidence.

\paragraph{Results.}
\Cref{tab:supp-overlap} reports each group.
GameScrub's removal does not rest on the matched clips: on the 30 identified clips that match no training title, edition or franchise it reaches 72.69 AAR and 91.73 UI, against 71.95 and 90.06 on all clips, and it leads GameCleaner (w.o. Ref) by 15.42 AAR points, with a 95\% interval of 2.4 to 28.0.
The matched clips are harder for both models: on the 16 same-title clips GameScrub obtains 54.36 AAR and GameCleaner (w.o. Ref) 48.57.
As on the full split (\cref{sec:supp-stats}), the differences in Overall are within their intervals in every group.


\subsection{Qualitative results by HUD type}
\label{sec:supp-qual}

\begin{figure*}[p]
  \centering
  \includegraphics[width=0.957\linewidth]{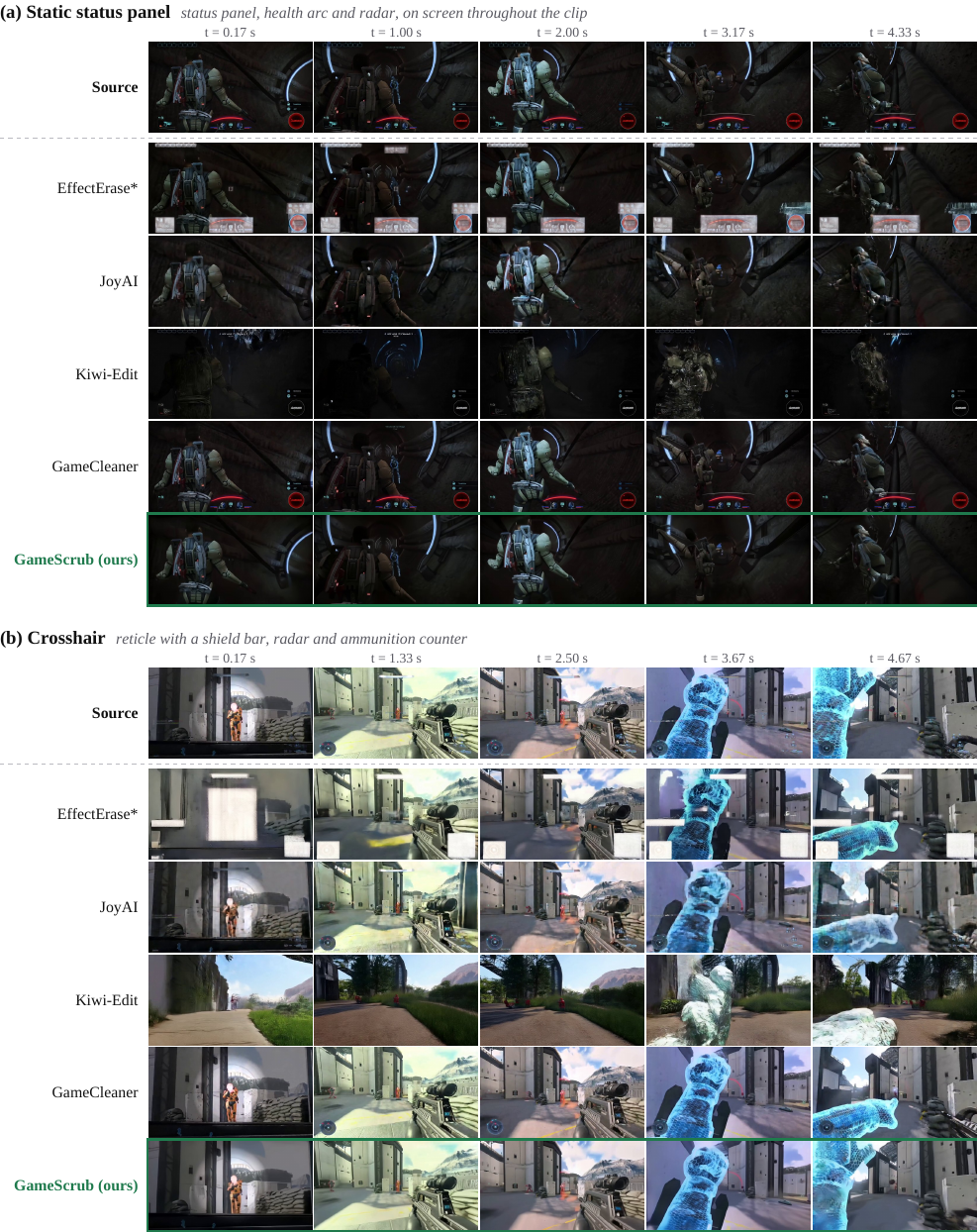}
  \caption{{\bf Screen-fixed HUD elements} on two in-the-wild clips. Rows give the source and the output of each method; columns give the same five moments of the clip. $^{*}$EffectErase is given the annotated boxes. GameCleaner is the standard configuration of \cref{tab:main}.}
  \label{fig:supp-hud-fixed}
\end{figure*}

\begin{figure*}[p]
  \centering
  \includegraphics[width=0.957\linewidth]{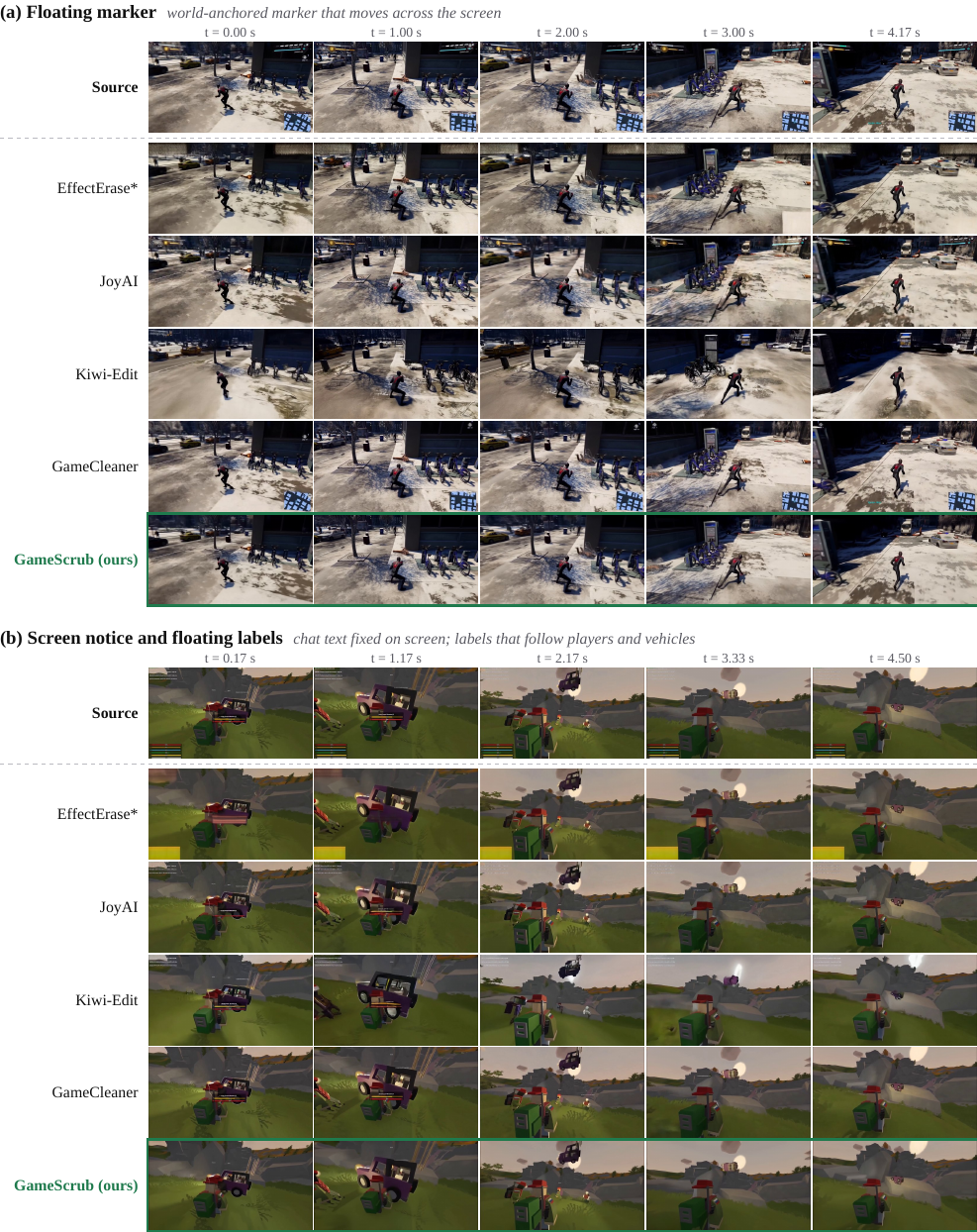}
  \caption{{\bf Moving HUD elements}: (a) a marker anchored to the scene; (b) labels that follow players and vehicles, next to chat text that stays fixed. Layout as in \cref{fig:supp-hud-fixed}.}
  \label{fig:supp-hud-moving}
\end{figure*}

\begin{figure*}[p]
  \centering
  \includegraphics[width=0.957\linewidth]{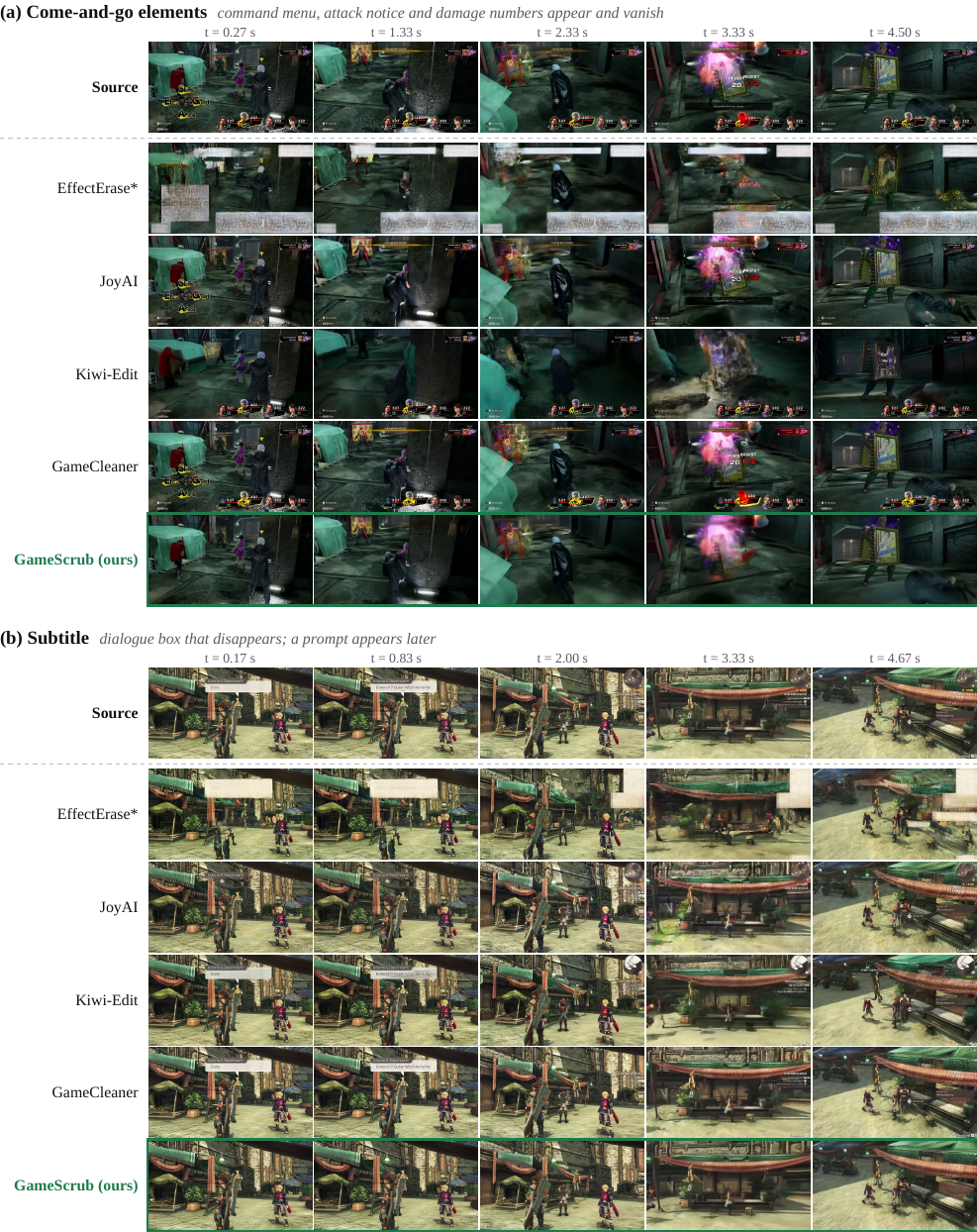}
  \caption{{\bf Transient HUD elements}: (a) a command menu, an attack notice and damage numbers that appear and vanish; (b) a dialogue box that disappears and an interaction prompt that appears later. Layout as in \cref{fig:supp-hud-fixed}.}
  \label{fig:supp-hud-transient}
\end{figure*}

The introduction distinguishes HUD elements that stay fixed on screen, elements that move across it, and elements that appear and vanish with game events.
\Cref{fig:supp-hud-fixed,fig:supp-hud-moving,fig:supp-hud-transient} show two in-the-wild benchmark clips for each kind; most clips also contain elements of the other kinds.
Each block gives the source clip and five outputs: EffectErase, which is given the annotated boxes; JoyAI and Kiwi-Edit, the two general editors with the highest AAR in \cref{tab:main}; GameCleaner in its standard configuration; and GameScrub.
The clips are chosen to illustrate each kind of element and are not a random sample; \cref{tab:main} reports the results on all 100 clips.

\paragraph{Frames.}
The columns of a block show the same five moments of the clip in every row.
The baseline outputs have 16\,fps and the source and GameScrub's output 30\,fps, so each baseline frame is the one nearest in time, at most 1/32\,s away.
All frames are cropped to the $1280{\times}704$ area common to every method.

\paragraph{Observations.}
EffectErase fills the boxes it is given with visible patches in most of the clips.
Kiwi-Edit changes the scene itself and replaces it entirely in \cref{fig:supp-hud-fixed}(b).
JoyAI and GameCleaner keep the scene but leave part of the HUD in place.
GameCleaner keeps the whole status panel in \cref{fig:supp-hud-fixed}(a), the minimap in \cref{fig:supp-hud-moving}(a) and the dialogue box in \cref{fig:supp-hud-transient}(b); JoyAI removes the panel of \cref{fig:supp-hud-fixed}(a) cleanly but keeps the radar in \cref{fig:supp-hud-fixed}(b) and the status bars in \cref{fig:supp-hud-moving}(a).
GameScrub removes the fixed panels, the moving markers and labels, and the transient notices in these clips.
Its outputs are not free of error: a faint trace of a name label remains at 4.67\,s in \cref{fig:supp-hud-transient}(b), and the region repainted over the damage numbers at 3.33\,s in \cref{fig:supp-hud-transient}(a) loses detail.
Such errors account for the gap between its UI and Clean scores in \cref{tab:main}.

\newpage
\bibliographystyle{plainnat}
\bibliography{references}

\end{document}